\def\year{2021}\relax
\documentclass[letterpaper]{article} 
\usepackage{aaai21}  
\usepackage{times}  
\usepackage{helvet} 
\usepackage{courier}  
\usepackage[hyphens]{url}  
\usepackage{graphicx} 
\usepackage{natbib}  
\usepackage{caption} 
\usepackage{subfig}
\usepackage{amsmath}
\usepackage{booktabs}
\title{Multi-object tracking with self-supervised associating network}
\author{

    Tae-young Chung, Heansung Lee, Myeong Ah Cho, Suhwan Cho, Sangyoun Lee*\\
    
}
\affiliations{
	Yonsei University\\
	\{tato0220,hslee2860,maycho0305,chosuhwan,syleee\}@yonsei.ac.kr

}

\begin{document}

\maketitle

\begin{abstract}
Multi-Object Tracking (MOT) is the task that has a lot of potential for development, and there are still many problems to be solved. In the traditional tracking by detection paradigm, There has been a lot of work on feature based object re-identification methods. However, this method has a lack of training data problem. For labeling multi-object tracking dataset, every detection in a video sequence need its location and IDs. Since assigning consecutive IDs to each detection in every sequence is a very labor-intensive task, current multi-object tracking dataset is not sufficient enough to train re-identification network. So in this paper, we propose a novel self-supervised learning method using a lot of short videos which has no human labeling, and improve the tracking performance through the re-identification network trained in the self-supervised manner to solve the lack of training data problem. Despite the re-identification network is trained in a self-supervised manner, it achieves the state-of-the-art performance of MOTA 62.0\% and IDF1 62.6\% on the MOT17 test benchmark. Furthermore, the performance is improved as much as learned with a large amount of data, it shows the potential of self-supervised method.

\end{abstract}

\section{1. Introduction}

Multi-Object Tracking(MOT) is one of the fundamental challenges of computer vision and is widely used in industrial fields such as surveillance image processing. many MOT studies have been conducted on pedestrians tracking, because the mainly used benchmark of MOT, the MOT challenge benchmark~\cite{milan2016mot16} is focused on pedestrians tracking.

In MOT, the detection by tracking paradigm has been dominated for a long time. It is a method of obtaining detection results in each frame of video using off-the-shelf detection algorithm such as Faster-RCNN~\cite{ren2015faster}, DPM~\cite{felzenszwalb2009object} , and organically associating the detection results of the previous frame with the detection results of the current frame and form tracks. The MOT algorithm can be divided into online method and batch method, which is the difference between whether future information is used to form tracks of the current frame. Online method that does not use future information is more suitable for practical application, a lot of research has been done with the online method.

There are various approaches in MOT algorithms, but the most of them deals with how to associate the track of the previous frame with the detection of the current frame. Among them, we approach by re-identifying detection results through appearance model. we train feature extraction network with appearance information of detection patches and associate tracks with detection patches using its feature.

However this approach has two drawbacks. First, a lot of computation is required because all detection patches have to pass through the feature extraction network and proceed matching with tracks using those features.
Second, the lack of learning data. Labeling MOT dataset is very labor intensive task. The MOT dataset consists of video sequence and a detection bounding box of each frame and its ID. To label this ID, a detection bounding box with an ID of all video sequence must be tracked and labeled by a human supervision. But each video sequence consists of many frames and also there are so many people per frame, it takes a lot of labor to label them. This is the one of the biggest problems in MOT.

Therefore, we try to solve this problem by applying self-supervised learning to MOT. Unsupervised learning or self-supervised learning has been proposed to solve the lack of training data problem in image classification or segmentation~\cite{doersch2015unsupervised,gidaris2018unsupervised,Vondrick_2018_ECCV}, but it has rarely been applied to MOT. We closely analyze the MOT to find a new pretext which is suitable for the MOT task and propose the self-supervised associating tracker(SSAT) which is a tracking algorithm that train the feature extraction network without data constraints by a self-supervised manner, and utilize it directly to re-identify targets to track without a seperate downstream tasks.

\section{2. Related work}
\subsubsection{Multi-object tracking} 
Since CNN-based object detectors have greatly improved detection performance, most MOT trackers have followed the tracking by detection paradigm. Main datasets such as the MOT challenge benchmark~\cite{milan2016mot16} and KITTI dataset~\cite{Geiger2012CVPR} provide detections as inputs, and gt is also composed of a continuous bounding box, so the main interest of MOT research has been how well associate the detections of two consecutive video frames. 

Motion-based trackers have been proposed. SORT~\cite{bewley2016simple} tracks bounding boxes using a Kalman filter and associates each bounding box with its highest overlapping detection. MOTDT~\cite{8486597} also uses Kalman filter to make a bounding box candidates and utilize with detection bounding box together. Then feature-based trackers such as DeepSORT~\cite{wojke2017simple} which applies appearance features from a deep network to SORT, idea of re-identifying target during tracking~\cite{kuo2011does} have also proposed.  OneShotDA~\cite{yoon2020oneshotda} use a network trained by one-shot learning to track. SiameseCNN~\cite{leal2016learning} use a siamese network to directly learn the costs between a pair of detections
 
However, Since feature-based trackers have drawbacks such as expensive computation and lack of training data so there are attempts to track without using features. 
predicting the track offset to associate detections~\cite{feichtenhofer2017detect} have proposed. Tracktor~\cite{bergmann2019tracking} associate detections with track using detector's box regression itself. CenterTrack~\cite{Zhou2020TrackingOA} predicts center detection heatmap and bbox size change and offset.

Nevertheless, the feature-based model is an effective method to improve tracking performance by retracking the missing track and is also robust in occlusion.
We propose a novel approach of feature-based MOT by solving the lack of training data, the one of the problems of feature-based models by a self-supervised method.

\subsubsection{Self-supervised learning}
One aspect of Self-supervised learning is pretext tasks. The pretext task means the task which is not for the real purpose, but a useful task to make network effective for the real purpose. A wide range of pretext tasks have been proposed such as denoising auto-encoders~\cite{vincent2008extracting}, context autoencoders ~\cite{pathak2016context}, or cross-channel auto-encoders (colorization)~\cite{zhang2016colorful,zhang2017split}. Some pretext tasks form pseudo-labels such as transformations of a single (“exemplar”) image~\cite{dosovitskiy2014discriminative}, patch orderings~\cite{doersch2015unsupervised,noroozi2016unsupervised}, visual object tracking~\cite{wang2015unsupervised} or segmenting objects~\cite{pathak2017learning} in videos, or clustering features~\cite{caron2018deep,caron2019unsupervised}.
In case of the real purpose of the self-supervised network is classification or segmentation, it is transferred to downstream tasks in a supervised manner. but since feature-based MOT only needs distinct features, self-supervised networks can be used without such downstream tasks. Our proposal extends the domain of self-supervised learning to MOT.

\section{3. Approach}
We try to solve the lack of training data problem in MOT by applying self-supervised learning. For our purpose, it is important to find a adequate pretext MOT learning, and we analyze the conditions of label for MOT feature learning as follows.

\begin{figure}[t]
	\centering
	\includegraphics[width=1.0\linewidth]{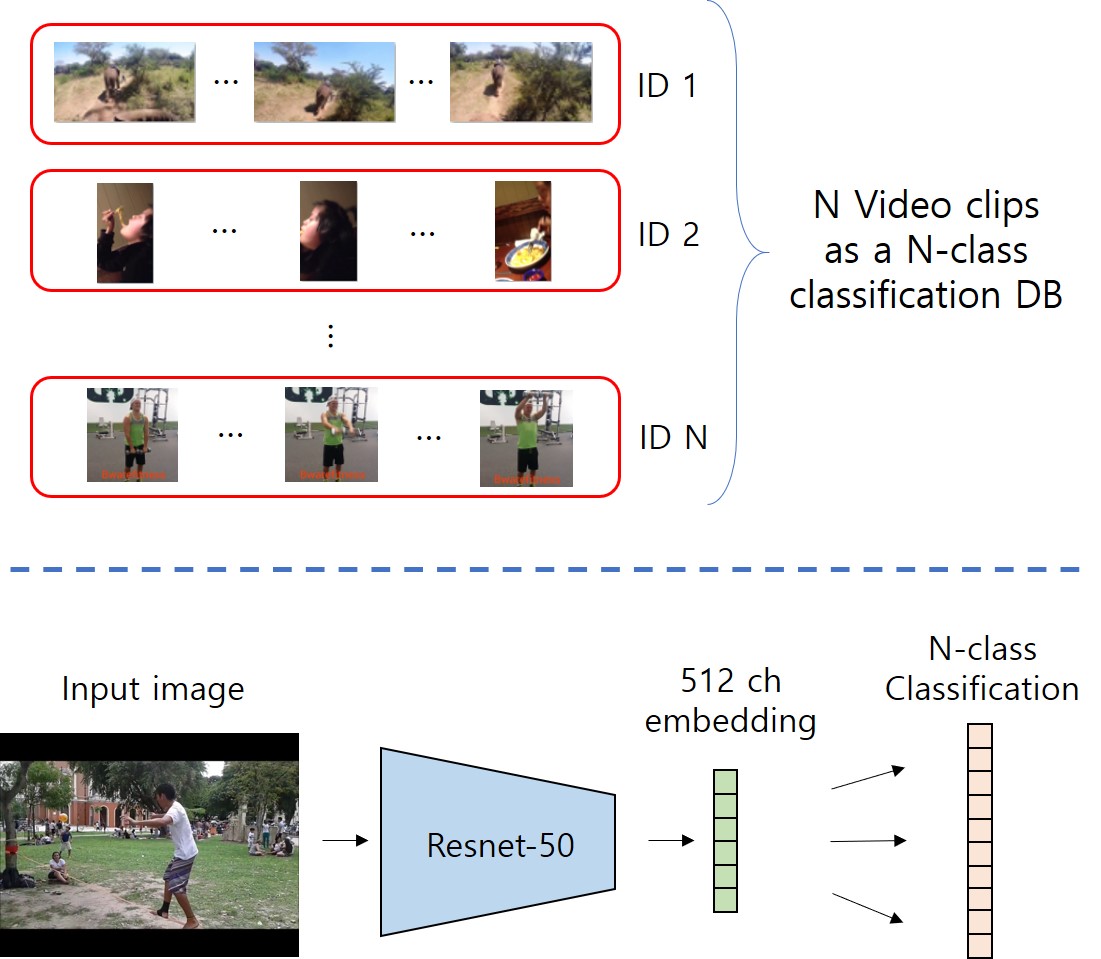}
	\caption{Main idea of our method. We consider all the frames of one short video as image patches with the same ID and train the network to classify N-class classification with N number video clips. Then we utilize output of the backbone network, 512 channel embedding as a feature of input image to associate detections with tracks.}
	\label{f1}
\end{figure}

\subsection{3.1 Conditions of label for MOT }
\subsubsection{Set of bounding boxes}
CNN is designed to utilize visual information, and if we train a feature extraction network based on CNN, the learning targets must be image patches in the form of bounding boxes. A dataset where several bounding boxes are classified according to their IDs is needed. 
\subsubsection{Continuous image patches through time}
Some MOT studies have been conducted to learn feature extraction networks using the person re-identification(re-ID) dataset~\cite{8486597}. Its' format is similar to MOT dataset, because it assign same ID to same pedestrian in various camera. So it can be additional dataset for supplement insufficient MOT dataset.
However, what the MOT algorithm actually requires from the feature extraction network is to find out how similar detection patches in the current frame is to the tracks of the previous frame. 
For the network to dealing with this problem, it is much more suitable to learn from continuous data through time. However, in the case of the person re-ID dataset, this temporal continuity cannot be guaranteed even among the targets of the same ID.
\subsubsection{Same ID taken from the same camera}
MOT’s robustness is not person re-ID's robustness. The person re-ID task is a task that finds the same person in pedestrian images captured in various camera environments. So person re-ID dataset assigns several pedestrian images at various brightness, angle, and resolution as the same ID. However, this is a little different from the capabilities required by MOT. In the MOT task, the feature which is robust to changes that can occur within one camera is required.


\subsection{3.2 Self-supervised feature extraction network}

MOT and person re-ID datasets~\cite{zheng2015scalable} assigned IDs for a large number of pedestrian patches by human supervision, and learned so that similar features appear between the same IDs. However, we think that the process of assigning an ID with the human supervision is main obstacle because it is very labor intensive task. So we decide to give up to train network with the pedestrian image. As figure~\ref{f1}, we train network with the frame itself as a patch. 

All frames that make up one video are considered as image patches with the same ID. it means that one video itself becomes a continuous patch bundle with the same ID. By this manner, it is possible to train with an infinite amount of data without any human supervision. And it meets the three conditions mentioned in section 3.1. 1) Although the target of a video is not pedestrian, but it consists of image patches with similar appearance. 2) Of course, it is a set of images that are continuous in time, and 3) all image is captured from one camera.

The conditions in section 3.1 are designed conditions to suitable for MOT, not Re-ID or classification, therefore the features obtained through these will be particularly suitable for MOT.

\subsubsection{10 seconds short video clip}
However, some videos may have a very large difference in appearance between frames due to editing or camera movement. In order to solve this problem, we assume that the frames of a sufficient short videos have similar appearance to each other,  use only short video clip to train the network. Some videos may be composed of completely different frames nonetheless, but we expected that if enough data is accumulated and learned, it will be solved by itself. Since self-supervised learning is free from labeling problems, it is possible to learn with a lot of data. thus, we set the short time to 10 seconds, secured a lot of short YouTube videos of about 10 seconds and learned this. In the case of a 30fps video, data in which 300 frames of images are assigned with one ID can be obtained.

\subsubsection{Kinetics dataset}
All the videos in this world were made with a intention. Even if it is a video of an autonomous vehicle or an installed surveillance camera, the intention of the person who caused the video to be filmed is reflected. We can train the network with any video, but we want to use the videos that would be suitable training for the feature extraction network for this MOT as a DB. Video of fixed landscape for a long time, or video with insane camera movement would not be a nutritious learning data no matter how short video is.

Thus we have found the DeepMind's Kinetics dataset~\cite{kay2017kinetics}. This kinetics-600 is a dataset created for 600 action recognition and has little to do with MOT. However, this dataset is full of short videos of about 10 seconds in which humans perform certain actions. This is exactly the dataset we are looking for. Some videos were too small or only part of a person appears(the hand of the person spinning disks, the hand grabing the hammer, etc.), but still each frames in a video have a similar appearance each others. In particular, the advantage of this dataset is that it's camera motion is focused on something in motion, because such a dataset is expected to contain more association between each frame. The Kinetics dataset provides a label of the action each video contains, but since it's not relevant to MOT task, we fetch only videos. 1600 random clips are selected and used for learning. The Kinetics dataset contains 500,000 clips, thus we can also learn with more data as we want

\begin{figure*}[h]
	\centering
	\includegraphics[width=0.80\linewidth]{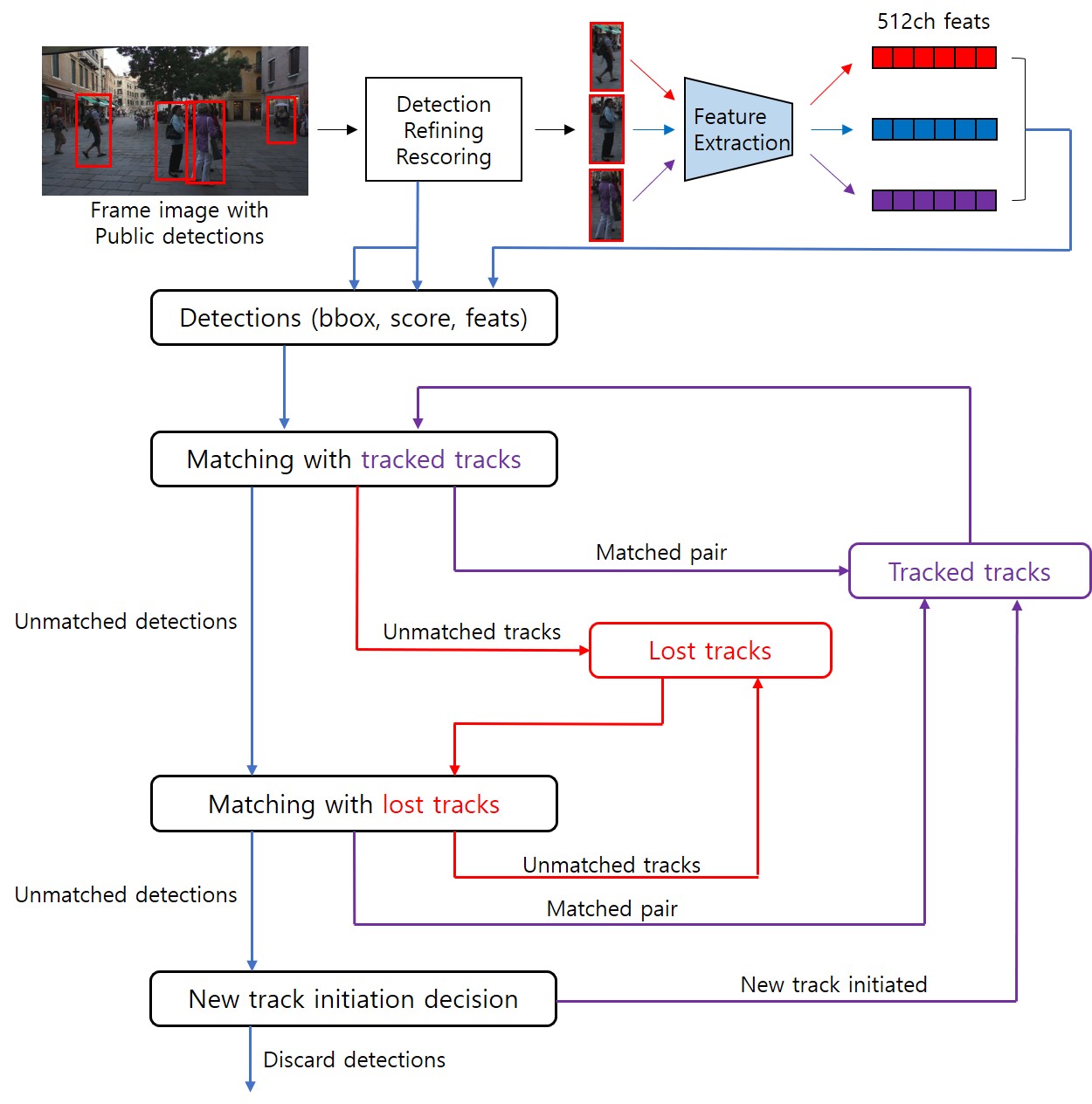}
	\caption{Overall framework of our tracker.}
	\label{f2}
\end{figure*}

\subsubsection{Training details}
Our network is simple to train. We assign each clip an ID as a label. And we train a classification network that outputs the ID of the video by taking a frame of the clip as an input.  We find that this process is very similar to face recognition task, thus we train the network by referring to Cosface~\cite{wang2018cosface}  which is known to be simple and effective in face recognition. Images used for training are resized to a minimum length of 256 pixels, then randomly cropped to 224x224 size patch and used as input. In general, when learning a pedestrian image, an input of 1: 2 or 1: 3 ratio is used, but the data we learn is not pedestrian-specific, and since it has various sizes and ratios, a 1:1 ratio is used. The backbone network is resnet-50, which extracts 512 channel features, then as the loss function, we use Large Margin Cosine Loss (LMCL) which is the loss function of CosFace. After training, when we apply our network to the MOT task, a feature of 512 channels which is the output of the backbone network is used to compare each patch with tracks.

\begin{table*}[h]
	\centering
	\begin{tabular}{lccccccc}
		\toprule
		Method & MOTA↑ & IDF1↑ & MT↑   & ML↓   & FP↓   & FN↓   & IDSW↓ \\
		\midrule
		MOTDT17~\cite{8486597} & 50.9  & 52.7  & 17.5  & 35.7  & 24069 & 250768 & 2474 \\
		YOONKJ17~\cite{yoon2020oneshotda} & 51.4  & 54    & 21.2  & 37.3  & 29051 & 243202 & 2118 \\
		FAMNet~\cite{chu2019famnet} & 52.0    & 48.7  & 19.1  & 33.4  & 14138 & 253616 & 3072 \\
		Tracktor++~\cite{bergmann2019tracking} & 53.5  & 52.3  & 19.5  & 36.6  & 12201 & 248047 & 2072 \\
		Tracktor++v2~\cite{bergmann2019tracking} & 56.3  & 55.1  & 21.1  & 35.3  & \textbf{8866} & 235449 & 1987 \\
		GSM\_Tracktor~\cite{liugsm} & 56.4  & 57.8  & 22.2  & 34.5  & 14379 & 230174 & \textbf{1485} \\
		CenterTrack~\cite{Zhou2020TrackingOA} & 61.5  & 59.6  & 26.4  & 31.9  & 14076 & 200672 & 2583 \\
		UnsuperTrack~\cite{karthik2020simple} & 61.7  & 58.1  & 27.2  & 32.4  & 16872 & \textbf{197632} & 1864 \\
		\midrule
		Ours(SSAT)  & \textbf{62.0} & \textbf{62.6} & \textbf{27.6} & \textbf{31.8} & 14970 & 197670 & 1850 \\
		\bottomrule
	\end{tabular}%
	\caption{Comparison of our method with recent online trackers in MOT17 benchmark. We achieve a state-of-the art performance.}
	\label{t1}%
\end{table*}%

\subsection{3.3 Overall tracking}
In the tracking by detection paradigm, most tracking algorithms can be divided into 2 steps. 1) Obtain detection results and in cases, refining or rescoring them. 2) Associating the detection result properly. Our work focuses on this second step and proposes a method of learning a re-identification network with self-supervised learning. However, detection refining in the first stage still requires the supervision datasets such as COCO datasets~\cite{lin2014microsoft}.

Thus, we developed a tracker based on CenterTrack~\cite{Zhou2020TrackingOA}, which performed well in MOT. CenterTrack uses CenterNet~\cite{zhou2019objects} to obtain or refine detection results and then associating results by its own method. In addition, UnsuperTrack, the recently proposed unsupervised method, is also based on CenterTrack, so we conduct ablation study with these three algorithms.

The overall tracking framework is as shown in the figure~\ref{f2}.
The video frame and its public detection are given in the input sequence. Then, the given detection is refined and rescored with CenterNet. And the detection results are passed through our network to obtain a 512 channel features. And if there is an existing tracked tracks, cosine similarity between the features of the tracks is computed, and if the smallest feature distance is less than certain threshold, it matches to the track and updates the feature and bounding box location to the track and continues tracking.

For each match, unmatched tracks or unmatched detections are occured. Unmatched tracks are identified as lost tracks and discarded if lost for a long time.
Unmatched detections proceed a second match with existing lost tracks. comparing is same as first match, but a more generous threshold is applied. If there is a matched pair, the lost track becomes a tracked track and continues to be tracked. Unmatched detections in the second match are regarded as the start of a new track, assign a new ID.
Furthermore, to improve tracking performance, we added two simple extensions in the matching steps.

\subsubsection{Comparing bounding box locations}
It is risky to matching with only appearance. Because two distant bounding boxes may have a similar appearance. to complement this problem, CenterTrack reflects the changes of the size and the center point of each bounding box. similar to this, we compute the difference between the each bounding boxes as a bounding box distance, $D_{bbox}$ by the equation~\ref{e1} and added it to the final matching distance.

\begin{gather}
	D_{bbox} = \frac{\left \| b_{track} - b_{detection} \right \|}{\alpha \cdot p_{track}}
	\label{e1}	
\end{gather}
where $b_{track}$ and $b_{detection}$ are 4 coordinates of bounding box in top-right bottom-left format and $p_{track}$ is perimeter of track bounding box. we use 2 as $\alpha$.\\

Then the final matching score $D_{final}$ computed as:
\begin{gather}
	D_{final} = D_{cos} + D_{bbox}
	\label{e2}
\end{gather}
where $D_{cos}$ is cosine similarity between features of detections and tracks.
Experimentally, ratio of $D_{cos}$ and $D_{bbox}$ is the best at 1:1.
we use the first matching threshold $\tau_1 = 0.56$ and the second matching threshold $\tau_1 = 0.64$ for more generous matching.

\subsubsection{Feature accumulation}
Whenever a feature is updated on a tracked track, replacing all of the features with a new feature is vulnerable to occlusion. Because once a feature is replaced with a distorted feature by occlusion, it is difficult to retrack. However, storing all past features of tracks is a very large waste of memory. Thus, we apply a weighted sum of features to updating features.

\begin{gather}
	f_{updated} = \beta \cdot f_{new} + (1-\beta)\cdot f_{track}
\end{gather}
where $\beta = 0.4$ is used.

\section{4. Experiment}
We submitted the results to the MOT Challenge benchmark~\cite{milan2016mot16} to evaluate the performance of our proposed tracker. The MOT17 dataset contains the MOT16 dataset, and many MOT algorithms have been submitted to the MOT17 benchmark, so we submitted it to the MOT17 benchmark. Although the feature extraction network learned in a self-supervised manner, our tracker acheived state-of-art performance.

\subsection{MOT Challenge benchmark}
The multi-object tracking benchmark, MOTChallenge, provides MOT data sets of 2D MOT 2015, MOT 16, MOT17 and MOT20. MOT20 has been released recently, but most of the algorithms have submitted to MOT17, which is suitable for comparison. MOT17 provides 7 test sequences and 7 training sequences for Pedestrian. In addition, for each sequence, a total of three types of public detections obtained by DPM~\cite{felzenszwalb2009object}, SDP~\cite{yang2016exploit}, and Faster RCNN~\cite{ren2015faster} are provided, and the performance of all public detections is evaluated together.

The main evaluation metrics are Multiple Object Tracking Accuracy (MOTA)~\cite{kasturi2008framework} and ID F1 score (IDF1)~\cite{ristani2016performance}. MOTA is a metric focused on object coverage. Since how much the bounding box is properly held is important in MOTA, well refined public detections tends to achieve high MOTA. IDF1 is a metric focused on identity preservation. Although the bounding box is similar to the ground truth, if the ID switch is frequent, you cannot get a good IDF1 score. MOT is evaluating performance by measuring metrics from these two perspectives together.

\subsection{Public detections}
The MOT benchmark recommends tracking using public detections provided by MOTChallenge to fairly compared to other tracking algorithms. However, refining and rescoring the given bounding boxes is allowed and is commonly used by participants in the challenge. Following Tracktor~\cite{bergmann2019tracking} and CenterTrack~\cite{Zhou2020TrackingOA}, we keep the bounding boxes that are close to an existing bounding box in the previous frame. New trajectory is only initialized if it is near a public detection.

\subsection{Implementation details}
As mentioned in Section 3.2, we extracted 1600 clips without labels from the Kinetics-600 testset and utilize them as training DB, and one clip consists of approximately 300 frames. Resnet-50 is used as the backbone network which output a 512 channel feature. And the classification of 1600 class clips is trained by applying the head of Cosface. We train for 16 epochs and lowered lr to 1/10 every 4 epochs from lr $10^{-2}$. We use the RTX2080, and a runtime of about 3.2fps come out on tracking.

\subsection{Benchmark evaluation}
Table~\ref{t1} is the results from the MOT17 test benchmark. The testset of the MOT challenge benchmark has not been public, the results must be submitted to the MOT Challenge official site to evaluate. We compared the algorithms that are published and online method among the algorithms submitted to MOT17 in table~\ref{t1}. our tracker ,SSAT achieves MOTA 62.0\% and IDF1 62.6\%.  In particular, IDF1 is relatively high, which can be seen that our algorithm was effective in identity preservation, and this shows that our tracker is efficient to associate detections.

\begin{table*}[h]
	\centering
	
	\begin{tabular}{lccccccc}
		\toprule
		\multicolumn{8}{c}{MOT17 testset.} \\
		\midrule
		Method & MOTA↑ & IDF1↑ & MT↑   & ML↓   & FP↓   & FN↓   & IDSW↓ \\
		\midrule
		UnsuperTrack~\cite{karthik2020simple} & 61.7  & 58.1  & 27.2  & 32.4  & 16872 & \textbf{197632} & 1864 \\
		CenterTrack~\cite{Zhou2020TrackingOA} & 61.5  & 59.6  & 26.4  & 31.9  & \textbf{14076} & 200672 & 2583 \\
		Ours(SSAT)  & \textbf{62.0} & \textbf{62.6} & \textbf{27.6} & \textbf{31.8} & 14970 & 197670 & \textbf{1850} \\
		\midrule
		\multicolumn{8}{c}{MOT17 trainvalset.} \\
		\midrule
		Method & MOTA↑ & IDF1↑ & MT↑   & ML↓   & FP↓   & FN↓   & IDSW↓ \\
		\midrule
		CenterTrack~\cite{Zhou2020TrackingOA} & 63.1  & 63.2  & 127   & 83    & 1555  & 17859 & 479 \\
		Ours(SSAT)  & \textbf{64.6} & \textbf{72.3} & \textbf{130} & \textbf{85} & \textbf{1403} & \textbf{17531} & \textbf{153} \\
		\bottomrule
	\end{tabular}%
	\caption{Comparison between tracker with same refining detections in MOT17 testset and trainvalset }
	\label{t2}%
\end{table*}%

\begin{table}[h]
	\centering
	\begin{tabular}{lccc}
		\toprule
		Method & MOTA↑ & IDF1↑ & IDSW↓ \\
		\midrule
		SSAT  & 63.7  & 68.5  & 256 \\
		+bbox loc & 64.4  & 72    & 160 \\
		+bbox loc +feature acc& \textbf{64.6} & \textbf{72.3} & \textbf{153} \\
		\bottomrule
	\end{tabular}%
	\caption{Ablation study of section 3.3's tracking extensions.\\ 
		bbox loc:   comparing bounding box locations.\\
	feature acc: feature accumulation.}
	\label{t3}%
\end{table}%

\begin{table}[h]
	\centering
	\begin{tabular}{lccc}
		\toprule
		number of training clip & MOTA↑ & IDF1↑ & IDSW↓ \\
		\midrule
		160 clips & 64.2  & 70.1  & 232 \\
		800 clips & 64    & 71.6  & 184 \\
		1600 clips & \textbf{64.6} & \textbf{72.3} & \textbf{153} \\
		\bottomrule
	\end{tabular}%
	\caption{performance difference by number of training clips. It shows that the more data, the better the performance. }
	\label{t4}%
\end{table}%

\section{5. Ablation study}
We conduct a few ablation study on the MOT17 trainval dataset to demonstrate and analyze effectiveness of our proposed tracker.

\subsection{Comparison between tracker with same refining detections}
CenterTrack, UnsuperTrack, and our proposal tracker apply the same refining detections method to public detections. Therefore, the difference between these three trackers is how associate the track and the detection bounding box. CenterTrack uses its network to estimating size and offset changes of bounding boxes, UnsuperTrack uses unsupervised trained feature extraction network learned with its own pseudo labeling.

Table~\ref{t2} is a comparison of the performance of three trackers. We find that our proposal tracker performs better than the other algorithms. In particular, it shows that change of the IDF1 score is larger than that of the MOTA. we analyze that the reason that the difference between the MOTA is small is that they use the same refining detection method, and ablity to associate tracks with detections make gap of IDF1 scores.

\subsection{Tracking extension}
In Section 3.3 we present two tracking extensions applied our tracker. Table~\ref{t3} is an ablation study for this. We find that combining spatial information with appearance model make quite performance improved. In some hyperparameters, the tracker achieves the higher MOTA, but there is a trade off that cause the IDF1 score dropped. Unfortunately, the effect of feature accumulation is insignificant.

\subsection{The more data, the better the performance}
We apply Self-supervised learning to MOT, assuming the more data makes the better performance. We experimentally checked our assumptions with trained network by 160 clips, 800 clips, 1600 clips in Kinetics DB. As Table~\ref{t4} shows, it is confirmed that the more data used, the higher the performance, that means the self-supervised learning has potential in MOT.

\section{Conclusion}
We have applied Self-supervised learning to MOT to solve the problem of lack of training data. We analyzed the conditions suitable for MOT and have proposed a novel pretext for self-supervised learning. We considered all the frames of one short video as image patches with the same ID, and trained the feature network without the limitation of labeled data. Through this, we acheive state-of-the-art performance in the MOTChallenge benchmark and demonstrate the validity of our proposal. We hope our work will be useful with other MOT and self-supervised learning study.

{
	\bibliographystyle{aaai}
	\bibliography{egbib}
}

\end{document}